\documentclass[preprint,12pt]{elsarticle}

\usepackage{amssymb}
\usepackage{amsmath}

\usepackage{multirow}
\usepackage{makecell}
\usepackage{subcaption}
\usepackage{tabularx}
\usepackage{hyperref}
\usepackage{inconsolata}
\usepackage{bm}
\usepackage{wrapfig}
\usepackage{arydshln}
\usepackage{hyperref}
\usepackage{url}
\usepackage{pifont}
\usepackage{color}

\journal{Neural Networks}

\newcommand{\cmark}{\ding{51}}%
\newcommand{\xmark}{\ding{55}}%

\begin{document}

\begin{frontmatter}





\title{SpikeCLIP: A Contrastive Language-Image Pretrained Spiking Neural Network}

\author[1]{Changze Lv\corref{equ}}
\ead{czlv24@m.fudan.edu.cn}
\author[1]{Tianlong Li\corref{equ}}
\ead{tlli22@m.fudan.edu.cn}
\author[1]{Wenhao Liu}
\author[2]{Yufei Gu}
\author[1]{Jianhan Xu}
\author[1]{Cenyuan Zhang}
\author[1]{Muling Wu}
\author[1]{Xiaoqing Zheng\corref{cor1}}
\ead{zhengxq@fudan.edu.cn}
\author[1]{Xuanjing Huang}

\affiliation[1]{organization={School of Computer Science, Fudan University},
            city={Shanghai},
            postcode={200433},
            country={China}}

\affiliation[2]{organization={University College London},
            city={London},
            country={UK}}
            
\cortext[equ]{Equal Contribution.}
\cortext[cor1]{Corresponding Author.}

\begin{abstract}
Spiking Neural Networks (SNNs) have emerged as a promising alternative to conventional Artificial Neural Networks (ANNs), demonstrating comparable performance in both visual and linguistic tasks while offering the advantage of improved energy efficiency.
Despite these advancements, the integration of linguistic and visual features into a unified representation through spike trains poses a significant challenge, and the application of SNNs to multimodal scenarios remains largely unexplored.
This paper presents SpikeCLIP, a novel framework designed to bridge the modality gap in spike-based computation. 
Our approach employs a two-step recipe: an ``alignment pre-training'' to align features across modalities, followed by a ``dual-loss fine-tuning'' to refine the model's performance. 
Extensive experiments reveal that SNNs achieve results on par with ANNs while substantially reducing energy consumption across various datasets commonly used for multimodal model evaluation.
Furthermore, SpikeCLIP maintains robust image classification capabilities, even when dealing with classes that fall outside predefined categories. 
This study marks a significant advancement in the development of energy-efficient and biologically plausible multimodal learning systems.
Our code is available at \url{https://github.com/Lvchangze/SpikeCLIP}.
\end{abstract}

\begin{graphicalabstract}
\begin{figure*}[]
\centering
\includegraphics[width=0.98\textwidth]{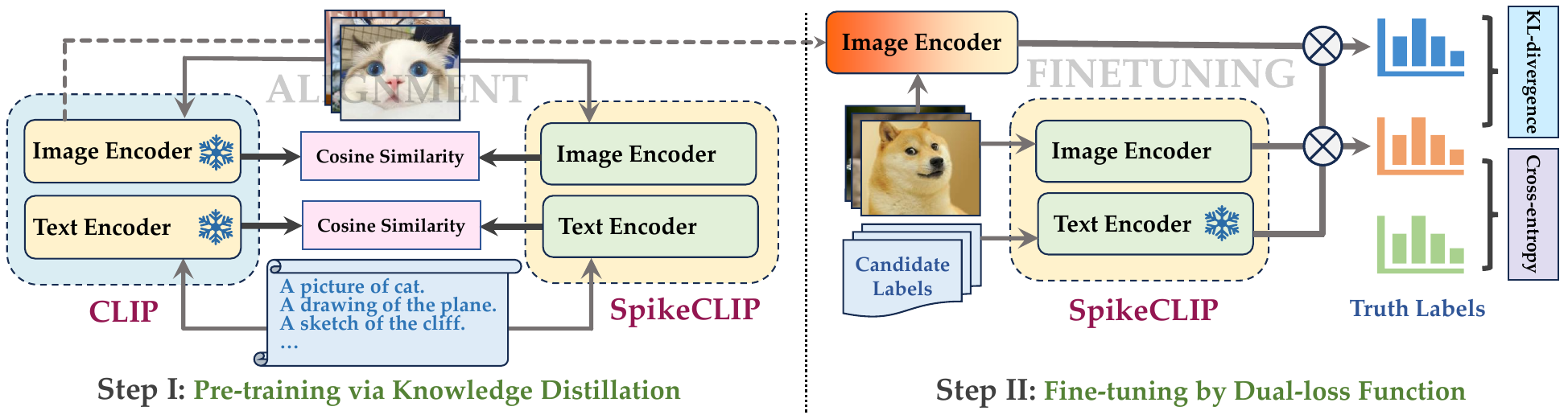}
\caption{The training framework of SpikeCLIP.}
\end{figure*}
\begin{figure*}[]
\centering
\includegraphics[width=0.95\textwidth]{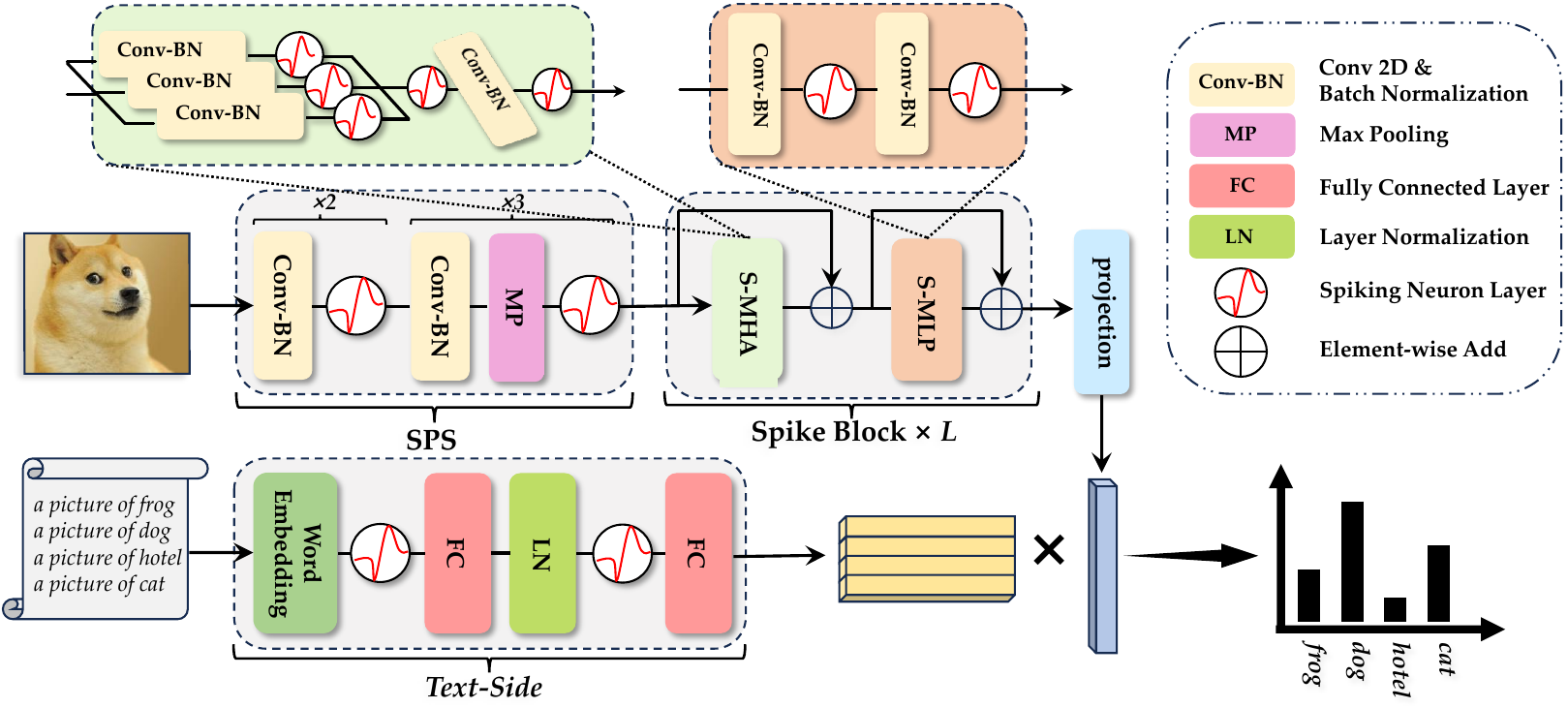}
\caption{The architecture of SpikeCLIP.}
\label{architecture}
\end{figure*}

\end{graphicalabstract}


\begin{highlights}
\setlength{\itemsep}{0pt}
\setlength{\parsep}{0pt}
\setlength{\parskip}{0pt}
\item {\color{black}\textbf{Spiking-Based Multimodal Feature Alignment}: This work is among the first to demonstrate that multimodal features extracted from text and images can be effectively aligned using spike train representations. These aligned representations enable zero-shot prediction of concept categories in previously unseen inputs.}
\item {\color{black}\textbf{Novel Training Algorithm for Multimodal SNNs}: We propose a two-step method for training multimodal SNNs, which includes pre-training for cross-modal alignment via knowledge distillation, followed by dual-loss fine-tuning with surrogate gradients.}
\item {\color{black}\textbf{Comprehensive Experimental Evaluation}: We conduct extensive experiments to assess the performance of SpikeCLIP on image classification tasks. Additionally, we perform ablation studies to demonstrate the model’s zero-shot capability and its reduction in energy consumption.}
\end{highlights}



\end{frontmatter}



\section{Introduction}

Artificial Neural Networks (ANNs) equipped with advanced deep learning techniques have demonstrated remarkable performance across a broad spectrum of visual and language tasks, sometimes even surpassing human capabilities \citep{krizhevsky2017imagenet,graves2014towards,mikolov2013distributed}.
However, the significant computational power and energy required to operate these cutting-edge deep neural models have been steadily escalating over the past decade. 
This substantial energy expenditure poses a major barrier to the widespread application of deep learning.
In contrast to ANNs, Spiking Neural Networks (SNNs) utilize discrete spikes for computation and information transmission, mirroring the energy efficiency of biological neurons. 
Neuromorphic hardware based on spike computation is now available and offers a more energy-efficient solution for implementing deep neural networks compared to specialized hardware such as GPUs.
It has been reported that improvements in energy consumption of up to $2 \backsim 3$ orders of magnitude when compared to conventional ANN acceleration on embedded hardware \citep{azghadi2020hardware,ceolini2020hand,davies2021advancing}.
Therefore, SNNs offer a promising computing paradigm to deal with large volumes of data using spike trains for information representation in a more energy-efficient manner.

{\color{black}
In reality, many neuromorphic systems experience performance loss during migration from simulation to hardware due to quantization or varying hardware support for operations, as shown in \cite{tang2021deep}.
}
However, mature on-chip training solutions are not yet available.
SNNs are typically trained using software simulation platforms, and then the resultant models are uploaded onto neuromorphic hardware for inference. 
Unlike their conventional ANN counterparts, it remains a great challenge to train SNNs due to the non-differentiability of discrete spikes, a challenge that persists even within software simulation environments.
{\color{black}
Recent intensive research \citep{cao2015spiking, diehl2015fast, rueckauer2017conversion, hu2018spiking, Yin2020EffectiveAE, Fang2021DeepRL, zhou2023spikingformer, zhou2023enhancing} on SNNs has significantly narrowed the performance gap between SNNs and ANNs, with this gap even disappearing in some vision tasks.
}
Although relatively fewer studies have explored the effectiveness of SNNs in Natural Language Processing (NLP) \citep{diehl2016conversion, rao2022long}, recent work indicates that spiking convolution networks can achieve results comparable to those of ANNs in multiple language datasets with significantly lower energy consumption \citep{lv2022spiking}. 
Despite these advances, existing research has predominantly focused on single-modality tasks, and the potential for applying SNNs to multimodal contexts remains largely unexplored.

\begin{figure*}[t]
\centering
\includegraphics[width=0.98\textwidth]{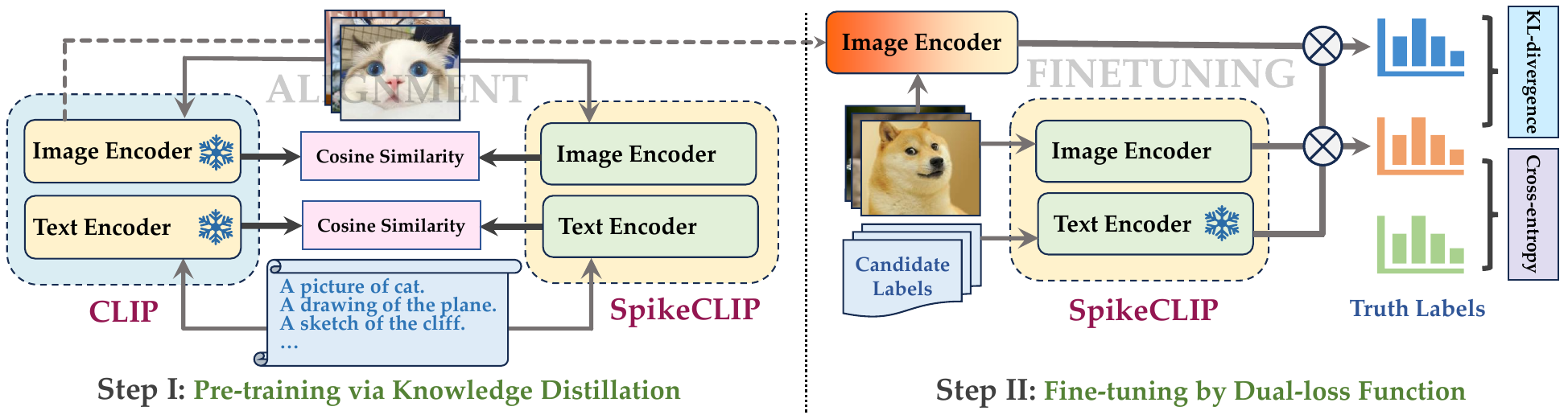}
\caption{ \label{fig:method}
\color{black}
An illustration of our two-step training approach for multimodal SNNs.
First, we pre-train SpikeCLIP by distilling knowledge from conventional CLIP, using a readout layer to map SNN states to floating-point feature representations.
Then, we fine-tune SpikeCLIP on downstream datasets, adding a regularization term based on Kullback-Leibler divergence on the training loss.
}
\end{figure*}

{\color{black}
A key challenge in extending the application of SNNs to multimodal contexts is the alignment of features extracted from multimodal inputs into spike train representations.
Achieving this alignment or mapping would enable the translation of texts and images into meaningful representations, which can then be used to assess the similarity of meaning across different modalities.
}
Moreover, if this mapping can make predictions about concept categories to unseen inputs, it would suggest that the underlying model has successfully captured meaning representations across modalities using discrete spikes. 
The approach to meaning representation that currently dominates the field of machine learning relies on distributed semantic representations, also known as embeddings.
One straightforward method for converting the spike trains produced by a collection of neurons into a representation is to interpret the firing rates as the activations of the representation \citep{cao2015spiking}.
With this conversion, cross-modal mapping can be achieved through contrastive learning, as demonstrated in the dual-stream CLIP \citep{radford2021learning}. 

Obtaining a high-performing cross-modal mapping through contrastive learning requires a substantial quantity of text-image pairs for the joint training of an image encoder and a text encoder.
It has been reported that as many as $400$ million pairs were used to train the CLIP.
Regrettably, these $400$ million text-image pairs are not yet publicly accessible.
Fortunately, the pre-trained CLIP has been made available to the research community, which enables us to employ the Knowledge Distillation (KD) technique \citep{hinton2015distilling} to train our spiking variant, which we have named SpikeCLIP.
However, distilling knowledge from ANNs and transferring it to SNNs is non-trivial, as there is no simple solution for representing negative values in SNNs.
To circumvent this problem, we used a readout layer to interpret the states of an SNN and map them to the feature representations that are amenable to knowledge distillation.
This approach draws on the principles of liquid state machines \citep{MAASS2004593,maass2011liquid}, a subclass of reservoir computer \citep{rohm2018multiplexed,tanaka2019recent} that uses SNNs for dynamic data processing.

{\color{black}
As shown in Figure \ref{fig:method}), after pre-training SpikeCLIP through the distillation of knowledge from the conventional CLIP, we proceed to fine-tune it on downstream datasets for image classification tasks.}
To enhance performance on instances whose classes fall outside predefined categories, we employ a dual-loss strategy that minimizes the cross-entropy between the predicted probabilities and actual distributions, while introducing a regularization term based on the Kullback-Leibler (KL) divergence.  
The purpose of this term is to impose penalties on any significant discrepancies between the feature representations generated by the SNNs and those produced by the CLIP model, which helps to maintain the generalization capacity gained during the pre-training stage.
Our ablation study demonstrated that such regularization can significantly enhance the performance of SpikeCLIP, particularly for images with categories not presented in the labels of a given dataset. 
To overcome the non-differentiable issue, we generalized the backpropagation algorithm with surrogate gradients \citep{zenke2021remarkable} to train the SNNs.

{
\color{black}
The contribution of this study can be summarized as follows:
\begin{itemize}
\setlength{\itemsep}{0pt}
\setlength{\parsep}{0pt}
\setlength{\parskip}{0pt}
\item \textbf{Spiking-Based Multimodal Feature Alignment}: This work is among the first to demonstrate that multimodal features extracted from text and images can be effectively aligned using spike train representations. These aligned representations enable zero-shot prediction of concept categories in previously unseen inputs.
\item \textbf{Novel Training Algorithm for Multimodal SNNs}: We propose a two-step method for training multimodal SNNs, which includes pre-training for cross-modal alignment via knowledge distillation, followed by dual-loss fine-tuning with surrogate gradients.
\item \textbf{Comprehensive Experimental Evaluation}: We conduct extensive experiments to assess the performance of SpikeCLIP on image classification tasks. Additionally, we perform ablation studies to demonstrate the model’s zero-shot capability and its reduction in energy consumption.
\end{itemize}
}

\section{Related Work}

\subsection{Training Methods for SNNs}

Spike neural networks have drawn considerable attention in recent years due to their potential to realize artificial intelligence while greatly reducing energy consumption.
Several training methods have been proposed to mitigate the non-differentiable of SNNs, which can generally be categorized into conversion-based and spike-based methods.
Conversion-based methods are to train a non-spiking network first and convert it into an SNN that inherits the learned weights of the non-spiking network \citep{cao2015spiking,diehl2015fast,sengupta2019going,rathi2020enabling,lv2022spiking}.
The advantage of such methods is that the non-differentiability of discrete spikes can be circumvented and the burden of training in the temporal domain is partially removed.
On the other hand, spike-based methods train SNNs using spike-timing information in either a supervised or unsupervised manner.

The majority of research \citep{hunsberger2015spiking,shrestha2018slayer,bellec2018long,huh2018gradient} in this line relies on the surrogate gradients training method, which estimated the back gradients with a differentiable approximate function so that gradient descent can be applied with backpropagation using spike times \citep{bohte2002error,booij2005gradient} or backpropagation using spikes (i.e., backpropagation through time).
{
\color{black}
\citet{gu2019stca,ma2023exploiting,wang2023adaptive} explore complementary aspects of SNNs, which contribute to a deeper understanding of spatiotemporal credit assignment, the role of noise in computation and learning, and adaptive gradient smoothing techniques.
}
These two training methods, or their combinations, have been investigated to train SNNs for single-modality tasks including computer vision or language processing.
In this paper, we propose a novel two-stage training method for multi-modal SNNs based on surrogate gradients.

{\color{black}
\subsection{SNNs for Single-Modal Tasks}
}

Many studies have shown that SNNs can yield competitive results in vision (mostly classification) tasks \citep{cao2015spiking,diehl2015fast,rueckauer2017conversion,shrestha2018slayer,sengupta2019going}.
\citet{cao2015spiking} pioneered a method that successfully converted a deep convolutional neural network into an SNN by interpreting the activations as firing rates. 
To minimize performance degradation during the conversion process, \citet{diehl2015fast} introduced a novel weight normalization method to regulate firing rates, which enhances the performance of SNNs in image classification tasks without additional training time.
\citet{sengupta2019going} pushed SNNs to go deeper by investigating residual architectures and introducing a layer-by-layer weight normalization method.
To mitigate performance loss after the conversion, 
\citet{bu2023optimal} suggested using a quantization clip-floor-shift activation function instead of the ReLU function in ANNs so that the spiking patterns of SNNs could be more accurately simulated during the training of the corresponding ANNs.
Inspired by the well-established Transformer architecture \citep{vaswani2017attention}, \citet{zhou2022spikformer} introduced a spiking version called Spikeformer, which was further refined to reduce reliance on floating-point computations \citep{zhou2023spikingformer,zhou2023enhancing}. By leveraging Transformer-like architectures, Spikeformer and its variants achieved state-of-the-art results in image classification tasks.

While the application of SNNs in the field of computer vision has been extensively investigated, their effectiveness in NLP tasks has been relatively less explored \citep{diehl2016conversion,rao2022long,lv2022spiking}. \citet{rao2022long} demonstrated that long-short-term memory (LSTM) units could be implemented on spike-based neuromorphic hardware using the spike frequency adaptation mechanism.
\citet{diehl2016conversion} used pre-trained word embeddings in their TrueNorth implementation of a recurrent neural network and achieved $74\%$ accuracy in a question classification task.
However, an external projection layer is required to project word embeddings to the vectors with positive values that can be further converted into spike trains.
\citet{lv2022spiking} proposed a two-step recipe of ``conversion + fine-tuning'' to train spiking neural networks for NLP. 
Initially, a normally trained ANN is converted into an SNN by duplicating its architecture and weights. Subsequently, the converted SNN undergoes fine-tuning. They showed that SNNs trained using this method can yield competitive results on both English and Chinese datasets compared to their ANN counterparts.
{\color{black}
\citet{lv2023spikebert,bal2023spikingbert} extended the Spiking Transformer \citep{zhou2022spikformer} to make it possible to process language tasks, resulting in the SpikeBERT. 
Their SNNs not only outperformed state-of-the-art models but also achieved results comparable to BERTs on certain text classification datasets while significantly reducing energy consumption.
}

{\color{black}
\subsection{SNNs for Multi-Modal Tasks}
Deep neural networks have shown their efficacy in multimodal modeling, which can be broadly categorized into single-stream architectures such as OSCAR \citep{li2020oscar} and SimVLM \citep{wang2021simvlm}, and dual-stream architectures like CLIP and WenLan \citep{huo2021wenlan}.
}
Single-stream models process multimodal inputs through a unified architecture, where all types of data are combined at an early stage before being fed into task-specific layers while dual-stream models process each modality through separate sub-networks first and then merge the outputs of these networks at a later stage.
Single-stream models are less flexible in handling the specific characteristics of each modality, as every type of input is treated uniformly through the same network layers.
In contrast, dual-stream models can optimize the processing for each modality independently, which can lead to better handling of the unique features of each data type.
Our preliminary attempts with single-stream architectures have not yielded the desired results probably because it is hard for SNNs to capture interactions between different types of data early in the forms of spike trains. 
However, we found that SNNs, when implemented with a dual-stream architecture and trained with a method that combines alignment pre-training with dual-loss fine-tuning, can rival the performance of their ANN counterparts in various multimodal classification tasks. Moreover, they display the capacity for zero-shot learning.

{\color{black}
Although the extensive exploration of SNNs in single-modality tasks, their potential application in multimodal contexts remains largely untapped.
Several recent studies \cite{liu2022event,guo2023transformer,jiang2023cmci} have demonstrated the feasibility of integrating multimodal information into SNN models. 
However, these studies primarily focus on modalities such as speech and images, with few addressing the fusion of text and image modalities.
\citet{panchev2005spiking} proposes a simple SNN module for robots to better understand language instructions; however, this module does not qualify as a multimodal SNN since it excludes the integration of both text and image data.
In contrast, our work is among the first to focus on the fusion of text and image modalities. We introduce a two-stage training method for multimodal SNNs and achieve promising performance across several benchmark datasets.
}

\section{Method}
 
\begin{figure*}[t]
\centering
\includegraphics[width=0.95\textwidth]{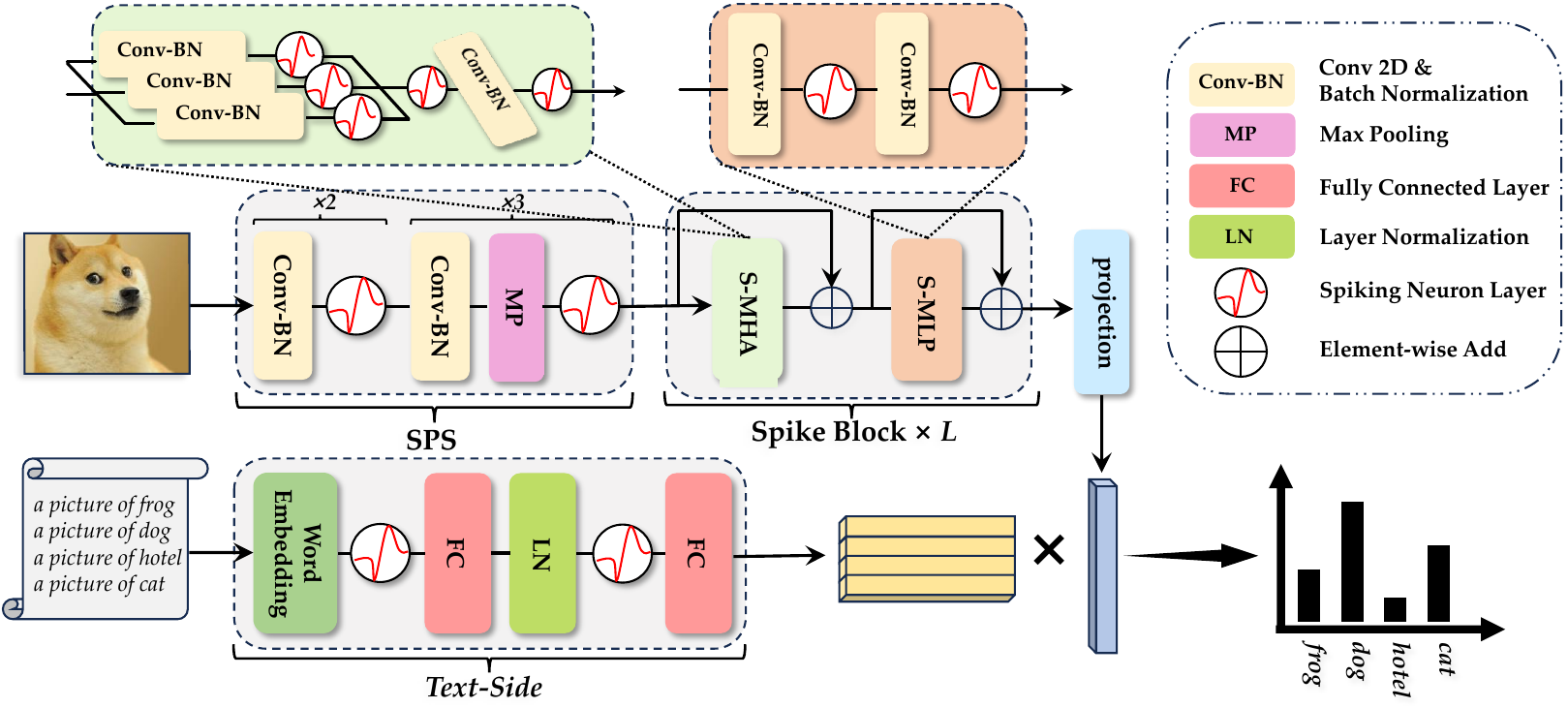}
\caption{
\color{black}
Overview of the architecture of SpikeCLIP.
We use dual-stream architectures for multimodal modeling.
The spiking image encoder is based on Spikingformer \citep{zhou2023spikingformer}, while a simple spiking MLP is designed for the text encoder of SpikeCLIP, with integrate-and-fire neurons converting data into spike trains for SNN processing.
}
\label{architecture}
\end{figure*}

{\color{black}
\subsection{Challenges and Motivations}
To enable the use of SNNs in image-text multimodal tasks, it is essential to align the semantic features extracted from both texts and images into spike train representations.
Given the limited flexibility of single-stream models in accommodating the unique characteristics of each modality, and the challenges of early-stage modality integration in SNNs, we adopt a dual-stream architecture for multimodal modeling.
This approach allows for more specialized processing of each modality before integration.

The successful paradigm of pre-training on large datasets to learn general features, followed by fine-tuning on task-specific datasets, has shown strong results in NLP and computer vision.
In line with this, we first pre-train SNNs by distilling knowledge from CLIP, then fine-tune the pre-trained networks on task-specific datasets using a dual-loss function.
We utilize a readout layer (see Section \ref{sec:pre-training}) to interpret the hidden states of the SNN, overcoming the challenge posed by the discrepancy between the floating-point feature representations of conventional ANNs and the temporal spiking representations of SNNs.

}

\subsection{Dual-Stream Architecture}

The dual-stream architecture was inspired by the human brain, which uses different regions to process different types of sensory information before integrating them for perception and decision-making.
In a dual-stream architecture, each ``stream'' or pathway is a sequence of layers that process a specific type of input data (see Figure \ref{architecture}). 
In our model designed to process both image and text data, one stream is dedicated to processing image data, while the other stream is used for text data. Each stream learns features independently from its specific type of data, and the features generated by one stream are compared with those produced by the other stream for prediction.

Given the remarkable success of the Transformer architecture \citep{vaswani2017attention}, \citet{zhou2023spikingformer} introduced a spiking variant, called Spikingformer, which achieved cutting-edge accuracy across multiple image classification datasets using event-driven spiking computations.
Therefore, we chose to employ Spikingformer in building the processing stream for image data.
To enable the conversion of spiking outputs into feature vector representations, a readout (MLP) layer was added on the top of the Spikingformer to build a full-fledged image encoder.
This layer also uses a set of learnable weights to integrate the spiking signal generated at different time steps (i.e. Time-Dependent Weight).
This approach is beyond the rate code solution, emphasizing the significance of the timing of emitted spikes.
\citet{lv2023spikebert} have shown that a relatively simple spiking convolutional neural network, bearing a similar architecture to TextCNN \citep{Kim2014TextCNN}, can deliver satisfactory accuracy across a range of datasets in both English and Chinese. 
{\color{black}
Given the shorter text lengths (up to 20 tokens) in the datasets used in this study compared to previous work, we chose a simpler architecture that employs a multi-layer perceptron (MLP) to construct the text encoder.
An ablation study of alternative architectural choices for text data processing can be found in Section \ref{sec:Text-side}.
}
Like the image encoder, a similar readout layer is also applied to transform spiking outputs into feature representations.

{\color{black}
\subsection{Building Block: Leaky Integrate-and-Fire Neuron}
}
\label{lif}
{\color{black}
Various spiking neuron models can be used to construct SNNs, and we chose the widely-used first-order leaky integrate-and-fire (LIF) neuron \cite{maass1997networks} as the foundational building block.
}
Analogous to traditional artificial neuron models, LIF neurons compute a weighted sum of inputs that contributes to the membrane potential $U_t$ of the neuron at time step $t$.
If this sum sufficiently causes the membrane potential to reach a predefined threshold $U_{\rm thr}$ and excites the neuron, the neuron emits a spike $S_t$:
\begin{equation} \label{eq:spike} \small
    \begin{split}
    S_t=
    \begin{cases}
    1, & \text{if  $U_t \geq$ $U_{\rm thr}$;} \\ 
    0, & \text{if  $U_t <$ $U_{\rm thr}$.} 
    \end{cases}
    \end{split}
\end{equation}

The dynamics of the neuron's membrane potential can be conceptualized as a resistor-capacitor circuit. 
An approximate solution to the corresponding differential equation for this circuit can be expressed as follows:
\begin{equation} \label{eq:potential} \small
\begin{split}
    U_{t} & = I_{t} + \beta U_{t-1} - S_{t-1} U_{\mathrm{thr}} \\
    I_{t} & = W X_{t}
\end{split}  
\end{equation}
\noindent where $X_t$ represents the inputs to the LIF neuron at time step $t$, while $W$ denotes a set of trainable weights that integrate these inputs. 
$I_{t}$ is the weighted sum of inputs. 
The parameter $\beta$ denotes the decay rate of the membrane potential, and $U_{t-1}$ is the membrane potential from the preceding time step $t-1$. 
The term $S_{t-1}U_{\mathrm{thr}}$ is introduced to account for the effects of spiking and the subsequent reset of the membrane potential.

\subsection{Converting Inputs into Spike Trains}

Spiking neural networks accept spike trains as inputs, and thus data from any modalities must be converted into spike trains for subsequent processing within SNNs.

For image data, we initially resized all images to a resolution of $224\times 224$ pixels. Each pixel in every channel can take a value within the range of $0$ to $255$.
These pixel values were then normalized by subtracting the mean of each pixel and dividing by its standard deviation.
{\color{black}
To convert the image data into spike trains, we employed direct encoding, a type of temporal encoding, by replicating the normalized pixel values for $T$ time steps.
In this process, the pixel intensity information is not explicitly transformed into spike timing but is instead maintained as a constant input over time.
This sustained input influences the membrane potential dynamics of spiking neurons, which in turn governs the temporal evolution of spike generation.
As a result, while direct encoding does not directly map pixel intensity to spike timing, it still falls under the category of temporal encoding since it leverages the temporal dimension to propagate information rather than encoding it in a single discrete step.
}
This procedure transforms the normalized pixel values into spike trains, with the corresponding spikes generated according to Equation \eqref{eq:spike}.
During the application of this Equation for the creation of spike trains, $W$ is consistently set to $1$, $X$ are normalized values, and $T$ is the number of time steps used for training SNNs.

{\color{black}
For text data, we adopt the approach of \cite{lv2023spikebert}, utilizing pre-trained word embeddings to enhance SNN performance.
In this method, a Poisson spike train is generated for each component of a word embedding, with the firing rate proportional to its value.
}
However, this approach requires a large number of time steps to accurately encode word embeddings, leading to increased energy consumption.
{\color{black}
Similarly, we use Equation \eqref{eq:spike} to directly transform the word embeddings after $T$ repetition into corresponding spike trains, where $X$ represents the values of the pre-trained embeddings.
}
Our empirical results show that this conversion is effective for multimodal modeling with SNNs.

\subsection{Pre-training SpikeCLIP via Knowledge Distillation} \label{sec:pre-training}

Leveraging the dual-stream architecture, SpikeCLIP is capable of generating feature representations for any given images (denoted as $x_\text{img}$) or texts (represented as $x_\text{txt}$) with the help of the readout layers.
Given a sufficient number of image-text pairs, cross-modal feature alignment can be achieved through contrastive learning.
However, as previously noted, there is an insufficiency of manually annotated image-text pairs available for the joint training of both the image and text encoders. 
To navigate this hurdle, we chose to use the knowledge distillation technique to pre-train our SpikeCLIP.
This approach not only enables the feasibility of pre-training for SNNs, but also substantially alleviates the difficulty in training these networks.

For pre-training the image encoder of SpikeCLIP, we used the ImageNet-1k dataset \citep{ILSVRC15}, which comprises approximately $1.28$ million images and is denoted as $\mathcal{D}_\text{img}$.
Following \citet{radford2021learning}, we applied a variety of prompt templates, such as ``\texttt{A photo of a \{label\}}'', over $1,000$ textual labels to generate nearly $116$ thousand sentences (denoted as $\mathcal{D}_\text{txt}$).
During the pre-training stage, we aimed to align the feature representations generated by SpikeCLIP with those produced by CLIP for both images and texts.
{
\color{black}
This was achieved by using the following loss function:
\begin{equation} \small
\label{Equation:Cosine-Loss}
\begin{split}
\small
\mathcal{L}_{\text{Pre-train}} \! = \! & \sum_{x_{\text{img}} \in \mathcal{D}_\text{img}} \left( 1 \! - \! \frac{E_c(x_{\text{img}}) \cdot E_s(x_{\text{img}})}{\|E_c(x_{\text{img}})\| \|E_s(x_{\text{img}})\|} \right) \! \\
 & + \! \sum_{x_{\text{txt}} \in \mathcal{D}_\text{txt}} \left( 1 \! - \! \frac{E_c(x_{\text{txt}}) \cdot E_s(x_{\text{txt}})}{\|E_c(x_{\text{txt}})\| \|E_s(x_{\text{txt}})\|} \right),
\end{split}
\end{equation}
}

\noindent where $E_c(\cdot)$ denotes the feature representation produced by CLIP, and $E_s(\cdot)$ denotes the feature vector generated by SpikeCLIP.
{\color{black}
The pre-training objective aims to maximize the cosine similarity between the two representations. While we used dual-stream architectures, we did not introduce separate notations for the image and text encoders, as the input modalities inherently distinguish them.
During pre-training, only the parameters of SpikeCLIP are updated by backpropagating errors through the layers to the word embeddings, with weight adjustments made using surrogate gradients.
}

\subsection{Fine-tuning Through Dual-loss Function} \label{sec:fine-tuning}

During the fine-tuning phase on a downstream dataset, the pre-trained SpikeCLIP generates feature representations for an input image and a set of possible labels using its image and text encoders.
{\color{black}
A prompt template is applied to each label to generate a sentence, which is then input into the spiking text encoder.
The image feature representation is subsequently used to compute the cosine similarity with each of the textual label representations.
}
These cosine similarity scores are subsequently converted into a probability distribution using a softmax operation.
Given an input image $x_{\text{img}}$, such a derived probability distribution, denoted as $y_{\text{img}}$, is compared to its true distribution $t_{\text{img}}$ (i.e., the ground truth label), and used to calculate the cross-entropy loss as follows:
\begin{equation} \small
\label{eq:cross-entropy}
\small
\mathcal{L}_{\text{CE}} = - \frac{1}{n} \sum_{i = 1}^n t_{
\text{img}} \log (y_{\text{img}})
\end{equation}
\noindent where $n$ is the number of training instances in a downstream dataset.
{\color{black}
Notably, only the weights of the image encoder are fine-tuned, while the text encoder weights remain fixed.
}
This strategy stems from the observation that the number of textual labels in a specific downstream dataset is significantly smaller compared to those explored during the pre-training phase. 
This strategy not only makes the training process more stable but also enhances the generalizability of the trained models to unseen labels.

To order to preserve the generalization capability obtained during the pre-training phase, we introduce a regularization term based on the Kullback-Leibler (KL) divergence. 
This term imposes a higher penalty when there is a substantial discrepancy between the probabilities predicted by SpikeCLIP and those predicted by CLIP.
This is equivalent to making the feature representations generated by SpikeCLIP not too far away from those produced by CLIP.
The KL-divergence is defined as follows:
\begin{equation} \small
\label{eq:KL-divergence}
\small
\mathcal{L}_{\text{KL}} = \frac{1}{n} \sum_{i = 1}^n h_{
\text{img}} \log \left(\frac{h_{\text{img}} + \epsilon}{y_{\text{img}} + \epsilon}\right)
\end{equation}
\noindent where $h_{\text{img}}$ denotes the probability distribution predicted by CLIP for an input image $x_{\text{img}}$, and a small constant $\epsilon$ is introduced to ensure numerical stability and prevent division by zero when calculating the KL-divergence ($\epsilon$ was set to $0.1 \times 10^{-9}$).
{\color{black}
The loss function employed during the fine-tuning stage is formulated as follows by integrating both the cross-entropy loss and KL-divergence term:
\begin{equation} \small
\label{eq:fine-tuning-loss}
\small
\mathcal{L}_{\text{FT}} = \lambda_1\mathcal{L}_{\text{KL}} + \lambda_2 \mathcal{L}_{\text{CE}}
\end{equation}
\noindent where the hyper-parameter $\lambda_1$ and $\lambda_2$ govern the relative importance of the regularization term compared with the cross-entropy loss.
}

\section{Experiments}

\begin{table*}[t]
\centering
\small
\caption{
\color{black}
Accuracy achieved on CIFAR10 and CIFAR100 datasets.
``Spike'' denotes whether the model is a SNN.
We report the modality type in the ``Type'' column.
The best and second-best results are highlighted in bold and underlined formats, respectively.
}
\resizebox{1.0\linewidth}{!}{
\begin{tabular}{l|c|ccc|cc}
\Xhline{1pt} 
\multirow{2}{*}{\bf Model} & \bf \multirow{2}{*}{Spike} & \bf \color{black} \multirow{2}{*}{Param. (M)} &\bf \multirow{2}{*}{Type} & \bf \multirow{2}{*}{Time Step} & \multicolumn{2}{c}{\bf Accuracy (\%)} \\
& & & & & \bf CIFAR10 & \bf CIFAR100 \\
\hline
\color{black} ViT \citep{dosovitskiy2010image} & \color{black} \xmark & \color{black} $86.39$ & \color{black} Unimodal & \color{black} -- & \color{black} $\bm{99.13}$ & \color{black} $\bm{94.20}$ \\
Hybrid Training \citep{rathi2020diet} & \cmark &\color{black}  $9.27$ & Unimodal & $125$ & $92.22$ & $67.87$ \\
Diet-SNN \citep{rathi2020diet} & \cmark & \color{black} $0.27$ & Unimodal & $10/5$ & $92.54$ & $64.07$ \\
STBP \citep{wu2018spatio} & \cmark & \color{black} $17.54$ & Unimodal & $12$ & $89.83$ & $--$ \\
STBP NeuNorm \citep{wu2018spatio} & \cmark & \color{black} $17.54$ & Unimodal & $12$ & $90.53$ & $--$ \\
TSSL-BP \citep{zhang2020temporal} & \cmark & \color{black} $17.54$ & Unimodal & $5$ & $91.41$ & $--$ \\
STBP-tdBN \citep{zheng2021going} & \cmark & \color{black} $12.63$ & Unimodal & $4$ & $92.92$ & $70.86$ \\
TET \citep{deng2022temporal} & \cmark & \color{black} $12.63$ & Unimodal & $4$ & $94.44$ & $74.47$ \\
Spikingformer \citep{zhou2023spikingformer} & \cmark & \color{black} $9.32$ & Unimodal & $4$ & $\underline{95.95}$ & $\underline{80.37}$ \\
\hline
\color{black} CLIP \citep{radford2021learning} & \color{black} \xmark & \color{black} $149.60$ & \color{black} Multimodal & \color{black} -- & \color{black} $\bm{98.45}$ & \color{black} $\bm{89.70}$ \\
SpikeCLIP (Ours) & \cmark & \color{black} $56.87$ & Multimodal  & $4$ & $\underline{94.48}$ & $\underline{77.69}$ \\
\hline
\Xhline{1pt} 
\end{tabular}
}
\label{tb:CIFAR}
\end{table*}

We conducted four sets of experiments.
{\color{black}
First, we evaluate the performance of SpikeCLIP against existing ANN and SNN baselines in image classification tasks.
Second, we assess the robustness of SpikeCLIP and its zero-shot learning capabilities across various image classification benchmark datasets.
Third, we investigate the impact of each component on SpikeCLIP’s performance and examine the choice of model architecture and key hyperparameters.
Finally, we compare the theoretical computing energy consumption of SpikeCLIP with that of its ANN counterparts.
For detailed implementation, datasets, and hyper-parameters, please refer to \ref{appendix:Datasets}, \ref{appendix:pretraining-data}, \ref{appendix:image-cls}, and \ref{appendix:ECR}.
}

\subsection{Image Classification} \label{sec:classification}

{\color{black}
We evaluated SpikeCLIP on two well-established CIFAR10 and CIFAR100 image classification datasets \citep{krizhevsky2009learning} against two ANN baselines, i.e., ViT \cite{radford2021learning} and CLIP \citep{radford2021learning}, and nine different spiking baselines, including Hybrid Training \citep{rathi2020enabling}, Diet-SNN \citep{rathi2020diet}, STBP \citep{wu2018spatio}, STBP NeuNorm \citep{wu2019direct}, TSSL-BP \citep{zhang2020temporal}, STBP-tdBN \citep{zheng2021going}, TET \citep{deng2022temporal}, TEBN \citep{duan2022temporal} and Spikingformer \citep{zhou2023enhancing}.
}
For each dataset, we trained all models on the training set and then evaluated them on the corresponding test set. 
We strictly adhered to the standard training and test splits as specified for each dataset.


{\color{black}
As shown in Table \ref{tb:CIFAR}, SpikeCLIP outperforms all SNN baselines, except for the unimodal Spikingformer.
}
Although Spikingformer achieved higher performance on these two datasets, the performance gap between SpikeCLIP and Spikingformer is relatively small with a difference of $1.47\%$ on CIFAR10 and $2.68\%$ on CIFAR100.
{\color{black}
In general, unimodal ANN models tend to perform better than their multimodal counterparts on specific datasets.
}
This is because multimodal models also possess the ability for zero-shot transfer to other datasets, albeit at the cost of some performance loss on a specific dataset.
For example, the conventional CLIP \citep{radford2021learning} achieved an accuracy of $98.45\%$ and $89.70\%$ on CIFAR10 and CIFER100 respectively, falling short of the performance of ViT \citep{dosovitskiy2010image} (one of the top-performing ANN unimodal models) by $0.68\%$ on CIFAR10 and $4.5\%$ on CIFAR100.
In comparison, the performance gap between SpikeCLIP and Spikingformer is less than that between CLIP and ViT with a difference of $0.515\%$ on average across these two datasets.
{\color{black}
These experimental results demonstrate that the performance discrepancy between multimodal SNNs and their unimodal counterparts can be narrower than that observed among ANNs.
}

\subsection{Zero-shot Results} \label{sec:zero-shot}

\begin{figure*}[t]
\small
\centering
\includegraphics[width=0.77\textwidth]{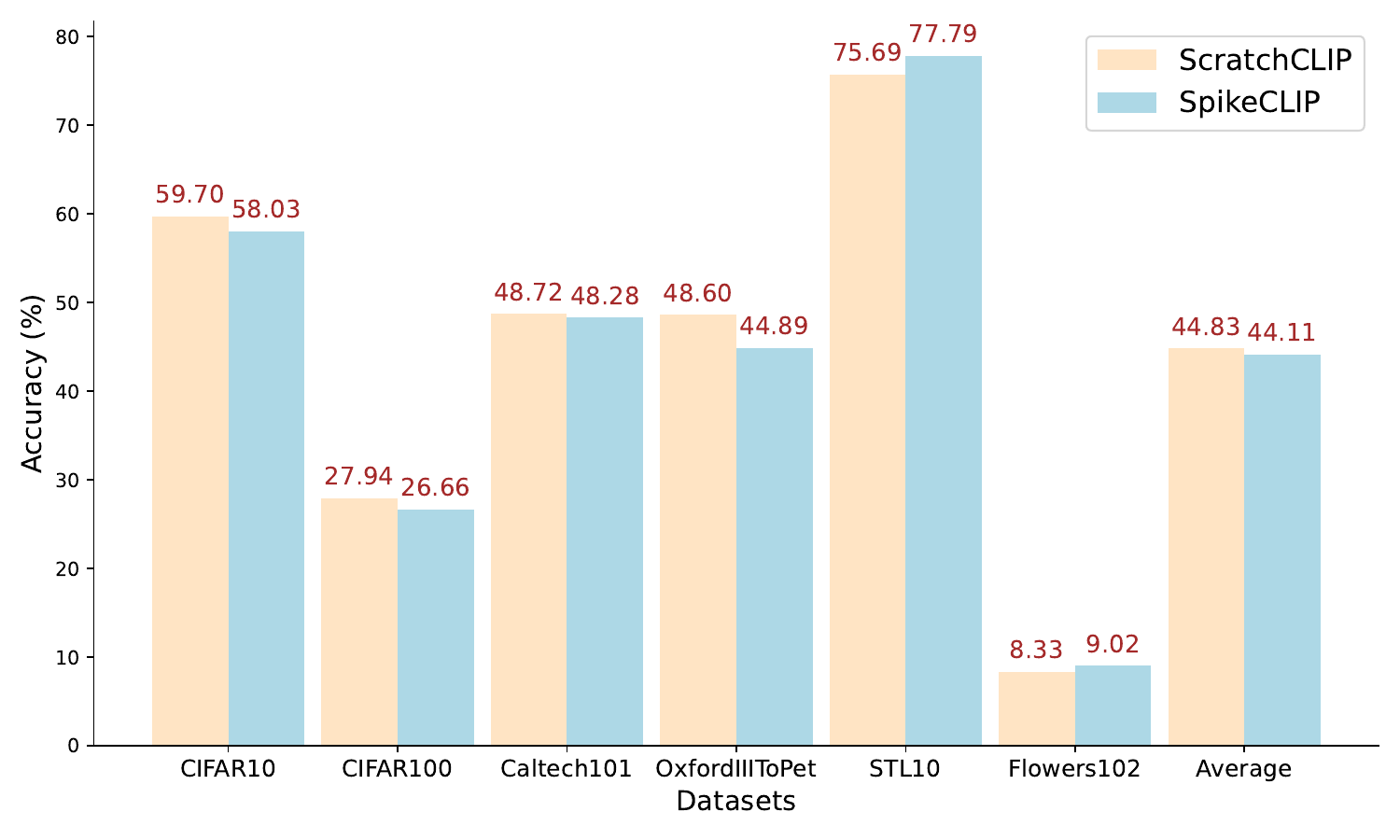}
\caption{\label{fig:zero-shot}
Zero-shot results on $6$ image classification datasets. {\color{black} SpikeCLIP achieved results comparable to ScratchCLIP, with a negligible average difference of only $0.72\%$.}
}
\end{figure*}
{
\color{black}
To the best of our knowledge, no previous spiking neural network has demonstrated zero-shot capabilities for image classification.
}
A direct comparison between SpikeCLIP and the conventional CLIP would be inappropriate, given that the latter was trained on $400$ million text-image pairs that are not yet publicly accessible.
Therefore, we constructed an ANN baseline, termed ScratchCLIP, which mirrors the architecture of SpikeCLIP but uses traditional artificial neurons instead of spiking ones.

To evaluate \textbf{zero-shot learning capabilities}, both ScratchCLIP and SpikeCLIP were only pre-trained by distilling knowledge from the CLIP without any following fine-tuning on downstream datasets.
ScratchCLIP was trained using the standard backpropagation algorithm.
{\color{black}
In addition to the CIFAR10 and CIFAR100 datasets, we follow the original CLIP setup and use 22 other datasets for zero-shot learning evaluation, including Caltech101 \citep{1384978}, OxfordIIITPet \citep{6248092}, STL10 \citep{DBLP:journals/jmlr/CoatesNL11}, Flowers102 \citep{DBLP:conf/icvgip/NilsbackZ08}, among others. Selected results are shown in Figure \ref{fig:zero-shot}, with detailed outcomes provided in \ref{app:zero26}.
}
Figure \ref{fig:zero-shot} illustrates that SpikeCLIP delivers results comparable to ScratchCLIP across $6$ different image classification datasets in a zero-shot setting. 
SpikeCLIP's performance was marginally inferior to ScratchCLIP, with a negligible difference of $0.72\%$ on average. 
This suggests that SNNs can potentially yield results that are on par with their ANN counterparts in multimodal zero-shot tasks despite the ease of training ANNs due to the absence of non-differentiability issues.

\begin{table*}[htp]
\centering
\small 
\caption{
{\color{black}
SpikeCLIP's accuracy on the CIFER10 and STL10 datasets under two challenging settings: first, label sets were enlarged by factors of $2$, $5$, and $8$ through noisy labels; second, textual labels were randomly replaced at rates of $20\%$, $40\%$, $80\%$, and $100\%$ with semantically equivalent expressions.
}
}
\resizebox{\textwidth}{!}{
\begin{tabular}{l|c|ccc|cccc}
\Xhline{1pt} 
\multirow{2}{*}{\textbf{Dataset}} &
\multirow{2}{*}{\textbf{Original}} &
\multicolumn{3}{c|}{\textbf{Introduction of Noisy Labels}} &
\multicolumn{4}{c}{\textbf{Replacement with Unseen Labels}} 
\\ \cline{3-9} 
& & \ \  $\times 2$ & \ \ \  $\times 5$ & 
$\times 8$ & $20\%$ & $40\%$ & $80\%$ & $100\%$  \\
\hline
CIFAR10 & $94.48$  & \ \  $94.48$ & \ \ \  $94.42$  & $94.38$  & $94.48$  & $94.48$  & $94.36$  & $94.27$    
\\
STL10  & $89.16$  & \ \  $88.85$  & \ \ \  $88.27$  & $87.95$ & $89.16 $  & $89.05$ & $88.18$  & $87.92$
\\ \Xhline{1pt} 
\end{tabular}
}    
\label{tb:robustness}
\end{table*}

{\color{black}
To assess the \textbf{robustness of SpikeCLIP}, we designed two additional, more challenging test scenarios.
}
The first involved the introduction of noisy labels that are semantically different from those in the original dataset's label set, yet difficult to differentiate from them. 
The second involved the replacement of certain portions of the textual labels with semantically equivalent expressions. 
In the first setting, we expanded the label sets by a factor of two ($\times 2$), five ($\times 5$) and eight ($\times 8$).
In the second setting, textual labels were randomly replaced at rates of $20\%$, $40\%$, $80\%$, and $100\%$.
For the detailed methods used to extend label sets and replace original labels, please refer to \ref{appendix:image-cls}.
{\color{black}
The empirical results presented in Table \ref{tb:robustness} indicate that SpikeCLIP consistently performed well under both challenging settings on the CIFAR10 and STL10 datasets, highlighting its robustness in handling noisy labels and previously unseen label variations.
}

\subsection{Ablation Study} \label{sec:Ablation}

{\color{black}
Firstly, to evaluate the impact of knowledge distillation during pre-training and the introduction of the KL-divergence term during fine-tuning on SpikeCLIP's performance, we conducted a series of ablation studies across six different datasets.
}
These studies involved eliminating the pre-training stage and removing the KL-divergence term from the loss function.

\begin{table*}[htp]
\centering
\small 
\setlength{\tabcolsep}{2.5pt} 
\caption{Empirical results from ablation studies.
These experiments were conducted by excluding the pre-training phase {\color{black}or fine-tuning phase (data from Figure \ref{fig:zero-shot})}, and removing the KL-divergence term from the loss function.}
\resizebox{\textwidth}{!}{
\begin{tabular}{l|cccccc|c}
\Xhline{1pt} 
\textbf{Method} &
\textbf{CIFAR10} &
\textbf{CIFAR100} &
\textbf{Flowers102} &
\textbf{Caltech101} &
\textbf{OxfordIIITPet} & 
\textbf{STL10} & \textbf{Average} \\ 
\hline

SpikeCLIP
& $94.48$  & $77.69$ & $86.07$  &  $82.31$ & $67.18$  & $89.48$ & $82.87$ \\ 

w/o Pre-training       
& $93.23$     & $74.59$     & $66.98$     & $23.67$     & $34.94$     & $69.25$     & $60.44$\\

\color{black} w/o Fine-tuning
& \color{black} $58.03$ & \color{black} $26.66$ & \color{black} $44.11$ & \color{black} $48.89$ & \color{black} $44.89$ & \color{black} $9.02$ & \color{black} $44.08$\\

w/o KL-divergence          
& $94.22$     & $77.52$     & $84.31$     & $79.74$     & $66.75$     & $65.29$     & $77.97$\\



\Xhline{1pt} 
\end{tabular}
}    
\label{tb:ablation}
\end{table*}

As indicated by the results presented in Table \ref{tb:ablation}, the full-fledged SpikeCLIP consistently achieved the highest accuracy across all evaluated datasets. Moreover, the incorporation of KD-based pre-training and the KL-divergence term enhanced the average performance by $22.45\%$ and $4.92\%$, respectively.
These findings indicate the critical role of pre-training via knowledge distillation in enabling SpikeCLIP to deliver performance on par with its ANN counterparts, and the use of KL-divergence significantly boosts SpikeCLIP's performance in a variety of image classification tasks. 

{
\color{black}
\begin{table}[htp]
\small
\centering
\caption{
\label{tab:lambda}
\color{black}
Performance on CIFAR10 and Flowers102 datasets for different $\lambda_1$ values.
Setting 1 refers to the scenario where, after fine-tuning in the second stage, the model is evaluated on the same dataset used for fine-tuning.
Setting 2 refers to the case where, after fine-tuning on one dataset, the model is evaluated on the other dataset.
}
\begin{tabular}{c|cc|cc}
\hline
\multirow{2}{*}{$\lambda_1$} & \multicolumn{2}{c|}{Setting 1} & \multicolumn{2}{c}{Setting 2} \\ \cline{2-5} 
& CIFAR10      & Flowers102     & CIFAR10      & Flowers102     \\ \hline
$0.5$ & $94.48$ & $86.88$ & $91.25$ & $74.52$ \\
$1.0$ & $94.48$ & $86.17$ & $90.85$ & $74.23$ \\
$2.0$ & $94.45$ & $86.25$ & $91.05$ & $74.15$\\
$10.0$ & $94.27$ & $86.08$ & $90.44$ & $73.52$ \\ \hline
\end{tabular}
\end{table}

Secondly, we also explore \textbf{the impact of the KL regularization term's coefficient $\lambda_1$ in the second stage}.
We conducted the hyper-parameter sensitive analysis of $\lambda_1$, as shown in Table \ref{tab:lambda}.
We perform experiments in two settings:
Setting 1 refers to the scenario where the second-stage fine-tuning is performed on a specific dataset, and the evaluation is carried out on the same dataset.
On the other hand, Setting 2 refers to the scenario where the second-stage fine-tuning is performed on one dataset, but evaluation is done on a different dataset.
We found that the value of $\lambda_1$ has minimal impact on the performance of SpikeCLIP during the second stage.
}

\begin{figure}[htp]
\centering
\begin{subfigure}[b]{0.45\textwidth}
\centering
\includegraphics[width=\textwidth]{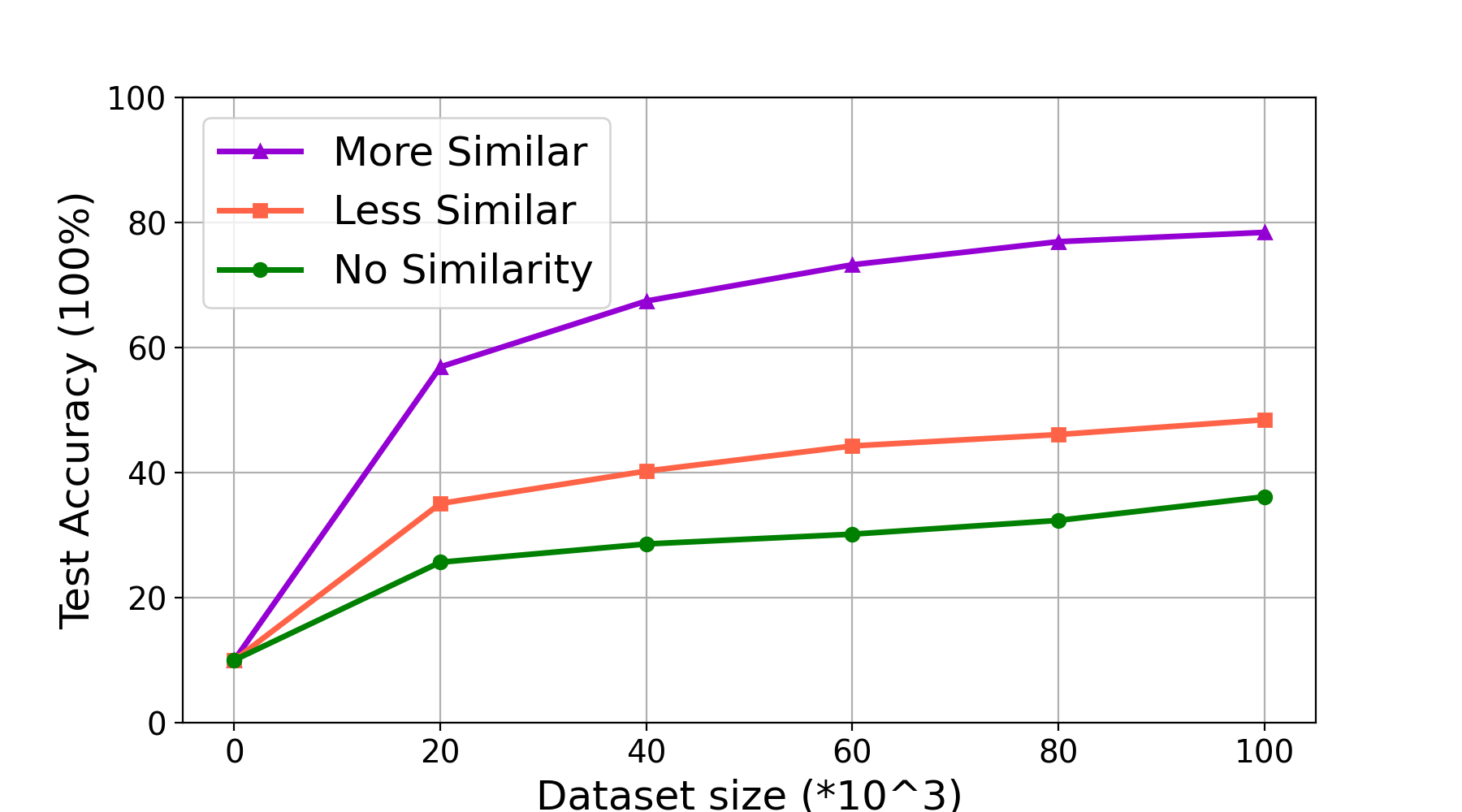}
\caption{Accuracy on CIFAR10 dataset.}
\end{subfigure}
\hfill
\begin{subfigure}[b]{0.45\textwidth}
\centering
\includegraphics[width=\textwidth]{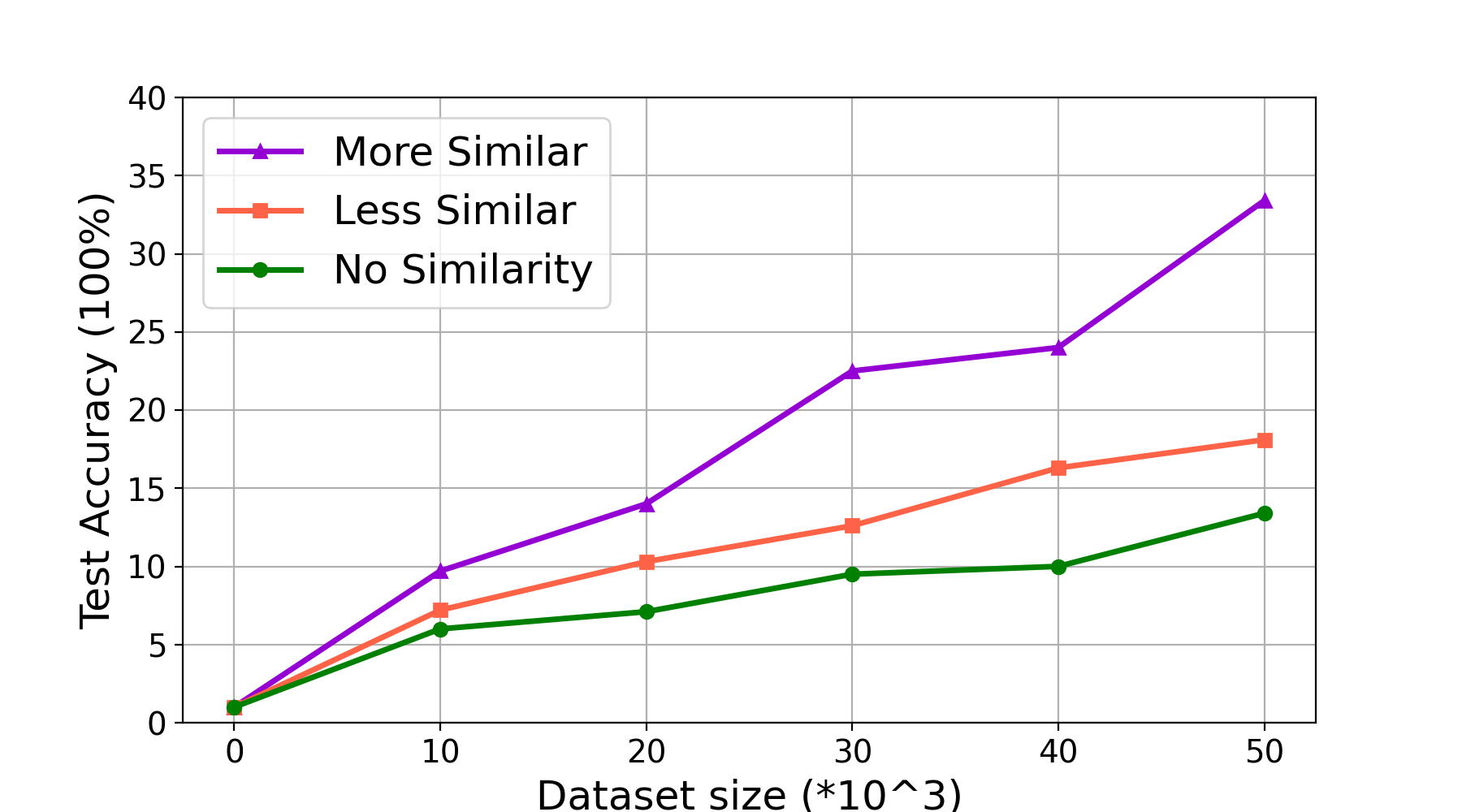}
\caption{Accuracy on ImageNet-1k dataset.}
\end{subfigure}
\caption{\label{fig:pretraining-data} Impact of the size and distribution of pre-training datasets on SpikeCLIP's performance.}
\end{figure}

Finally, we aim to investigate {\color{black}\textbf{how the size and data distributions of the pre-training datasets influence SpikeCLIP's performance.}
}
In addition to investigating the impact of diverse dataset sizes, {\color{black} we also explored how the degree of overlap between pre-training data and downstream task data influences performance (Figure \ref{fig:pretraining-data}).}
We evaluated three different data distributions during the pre-training phase: one included one-third of the downstream task data (indicated by ``More similar''), another excluded downstream task data completely (indicated by ``Less similar''), and the third represented a scenario falling between these two extremes (indicated by ``No similarity'').
{\color{black}
Further details on the creation of these datasets are provided in \ref{appendix:pretraining-data}.
}
As hypothesized, a direct correlation is observed between the growth in the size of pre-training data and the usage of more downstream task data during the pre-training phase.
These experimental findings suggest the potential for further enhancing SpikeCLIP's performance by expanding the size and coverage of the pre-training dataset.

{\color{black}
\subsection{Impact of Text Encoder Architectures}
\label{sec:Text-side}

To draw a comparison with the CLIP model, we initially employed Transformer-based architecture for both the text and image encoders, trained on the dataset \textit{D-text} constructed in this study. However, the Transformer-based text encoder struggled with effective loss minimization during training and suffered from poor accuracy when integrated with the image encoder. 

\begin{table}[htp]
\begin{center}
\small
\setlength{\tabcolsep}{2pt}
\caption{Comparative analysis of two network architectures used as text encoders across six text classification benchmarks.}
\label{appendix-tab:text-side}
\resizebox{\textwidth}{!}{
\begin{tabular}{l|cccccc|c}
\Xhline{1pt} 
\textbf{Architecture} & \textbf{CIFAR 10} & \textbf{CIFAR 100} & \textbf{Caltech 101} & \textbf{Flowers 102} & \textbf{OxfordIIITPet} & \textbf{STL 10} & \textbf{Average} \\
\hline
Transformer-based & $86.37$ & $48.03$ & $75.78$ & $27.09$ & $33.93$ & $94.76$  & $60.99$ \\
MLP-based &  $90.63$ &  $64.69$ &  $79.88$ &  $62.86$ &  $81.79$ &  $97.58$ &  $79.57$  \\
\Xhline{1pt} 
\end{tabular}
}    
\end{center}
\end{table}

An improvement was noted upon switching to a Multi-Layer Perceptron (MLP-based) architecture for the text encoder, by following the work \cite{bal2023spikingbert}. Our observation suggested that within the two-step training scheme (Pre-training + Fine-tuning), the text encoder is prone to overfitting if the architecture is overly complex and the architecture of MLP is proven to be proficient in this study. Comprehensive experimental results are presented in Table \ref{appendix-tab:text-side}.
}

{\color{black}
\subsection{Impact of Learnable Time-dependent Weights on Spiking Integration}
\label{sec:TSW}

In previous SNNs, tensor values were averaged across different time steps ($T$) before being classified. 
However, this approach assigns the same weight to each step ($1/T$), ignoring their interdependence. In particular, if the previous time step has already produced a spike, it may be more difficult for the current time step to produce a new spike again, so the signal from the new spike generated by the current time step may be stronger. 

This idea is not considered in cases where different time steps are given the same weight, which can lead to reduced performance. 
To address this issue, we employ learnable parameters to replace the fixed averaging weights, which are incorporated into SpikeCLIP.

\begin{table}[htp]
\centering
\small
\setlength{\tabcolsep}{10pt}
\caption{The impact of learnable time-dependent weights on model's performance.}
\label{Appendix-tab:TSW}
\begin{tabular}{l|cccc}
\Xhline{1pt}
\textbf{Dataset} & \textbf{Baseline} & \textbf{AD} & \textbf{AR} & \textbf{TDW} \\ 
\hline
CIFAR10  & $94.39$ & $94.39$ & $94.45$ & $\bm{94.48}$ \\
CIFAR100 & $77.51$ & $77.58$ & $77.56$ & $\bm{77.69}$ \\
\Xhline{1pt}
\end{tabular}
\end{table}

For benchmarking purposes, we also examined two sets of fixed parameters: one based on arithmetic differences (\textbf{AD}) and another based on arithmetic ratios (\textbf{AR}). 
Experimental outcomes corroborate the efficacy of our proposed Time-Dependent Weight (\textbf{TDW}) mechanism (As shown in Table \ref{Appendix-tab:TSW}).
}

\subsection{Comparison of {\color{black}Computing} Energy Consumption}
\label{sec:Energy}

\begin{table*}[htp]
\centering
\small 
\setlength{\tabcolsep}{3pt}
\caption{
\color{black}
Estimation of computing energy consumption on six image classification benchmarks. The application of SpikeCLIP results in an average energy reduction of approximately $78\%$.}
\resizebox{\textwidth}{!}{
\begin{tabular}{l|cccccc}
\Xhline{1pt} 
\textbf{Dataset} & \textbf{CIFAR10} & \textbf{CIFAR100} & \textbf{Flowers 102} & \textbf{Caltech101} & \textbf{OxfordIIIPet} & \textbf{STL10} \\ \hline
Firing Rate (\%)      & $27.26$   & $28.98$    & $29.30$      & $27.97$      & $27.93$        & $27.56$ \\ 
Energy Consumption (mJ)            & $3.17$    & $3.37$     & $3.41$       & $3.25$       & $3.25$         & $3.21$  \\ 
Energy Reduction Rate (\%) & $78.66\downarrow$   & $77.31\downarrow$    & $77.06\downarrow$      & $78.10\downarrow$      & $78.13\downarrow$  & $78.42\downarrow$ \\ 
\Xhline{1pt} 
\end{tabular}
}
\label{Table:Energy_consumption}
\end{table*}

{\color{black}
We follow \citet{yao2022attention,zhou2022spikformer} to conduct an analysis on estimating the \textbf{computing theoretical energy consumption} of SpikeCLIP across six distinct image classification datasets and reported the results in Table \ref{Table:Energy_consumption}.
}
The way to calculate the firing rate ($\%$), energy consumption (mJ), and energy reduction rate ($\%$) can be found in \ref{appendix:ECR}. 
As we can see from Table \ref{Table:Energy_consumption}, SpikeCLIP can achieve an average {\color{black}computing} energy consumption reduction of approximately $78\%$ on average.
This significant reduction is attributed to the sparse activation of its neurons (i.e., not operating at $100\%$ firing rates) and the event-driven nature of the inferences.

{\color{black}
Most importantly, we would like to clarify that our energy consumption estimation focuses solely on computing energy, excluding factors such as memory access and data movement.
A detailed discussion of this limitation can be found in \ref{app:limit}.
}

\section{Conclusion and Future Work}
\textbf{Conclusion} \quad
We found it hard to train SNNs for multimodal tasks directly due to the challenge of integrating linguistic and visual features into a unified representation through spike trains.
To circumvent this obstacle, we suggested a two-step training recipe: an initial phase of pre-training for cross-modal alignment via knowledge distillation, followed by dual-loss fine-tuning using surrogate gradients.
{
\color{black}
A readout mechanism was proposed to interpret the states of SNNs to enable knowledge distillation from ANNs, and a regularization term was introduced to preserve the generalization capacity attained during the pre-training phase.
}
Through extensive experimentation on $6$ image classification datasets, we demonstrated that the SNNs trained with the proposed method can match the performance of their ANN counterparts in multimodal classification tasks and exhibit zero-shot learning capabilities.

{
\color{black}
\textbf{Future Work} \quad
The following are our plans for scaling up to larger datasets in future work to improve generalization capabilities, particularly for zero-shot tasks:
First, it is indeed challenging to obtain a dataset of the same scale as CLIP’s full training dataset, as you said.
However, with the rapid development of large multimodal models, we anticipate that access to larger synthetic multimodal datasets, such as LAION-5B \cite{schuhmann2022laion}, will become feasible in the future.
Then, we will follow the experimental settings of CLIP \cite{radford2021learning} and conduct zero-shot experiments after the pre-training on the large datasets.
Secondly, the primary objective of this work was to explore the feasibility of modality fusion between text and images using a spiking neural network (SNN) architecture.
Our experiments demonstrate that this approach is viable and promising, providing a potential pathway to reduce the energy consumption of future multimodal large models.
Finally, we note that in biological systems, multimodal signals (such as sound, images, and speech) are processed using spike signals.
Our work validates the biological plausibility of integrating multimodal information using spiking neural networks.
We believe that these insights can contribute to the future development of more efficient multimodal models.
Limitations are discussed in \ref{app:limit}.
}

\section*{Broader Impact}
\label{sec:broader_impact}

The goal of this research is to propel advancements in the domain of Spiking Neural Networks (SNNs).  While conventional artificial neural networks (ANNs) have found extensive practical applications, SNNs remain predominantly within the realm of fundamental exploration. 
As per our assessment, this work is not anticipated to engender any negative societal implications.

\section*{Reproducibility Statement}

The authors have made great efforts to ensure the reproducibility of the empirical results reported in this paper. 
{
\color{black}
To begin with, the experiment settings, evaluation metrics, and datasets were described in detail in Section \ref{sec:classification}, Section \ref{sec:zero-shot}, and \ref{appendix:Datasets}, \ref{appendix:pretraining-data}, \ref{appendix:ECR}.
Furthermore, the implementation details were clearly presented in Section \ref{sec:pre-training}, Section \ref{sec:fine-tuning} and \ref{appendix:main-experiment}, Section \ref{sec:TSW}. 
}
In our effort to ensure reproducibility, we have submitted the source code of the proposed training algorithm with our paper, and plan to release the source code on GitHub upon acceptance.

\section*{Acknowledgements}
The authors would like to thank the anonymous reviewers for their valuable comments. This work was supported by National Natural Science Foundation of China (No. 62076068).

\clearpage

\appendix

\section{Implementation Details of Training Method}
\label{appendix:main-experiment}
For the pre-trained CLIP model, we use \href{https://huggingface.co/openai/clip-vit-base-patch16}{openai/clip-vit-base-patch16} with a dimension of 512 in this study. A Spikingformer-4-384 (\citep{zhou2023enhancing}) with 4 layers and a dimension of 384 is used as the base model for comparison. The image-side component architecture of SpikeCLIP is built upon this base model with a time-step weight (TSW) layer followed by a dimensionality-mapping layer, aligning the output to a 512-dimensional space compatible with pre-trained CLIP models.

For comparing purposes with SpikeCLIP, we constructed ScratchCLIP as an ANN counterpart. The image encoder of ScratchCLIP has a 4-layer Transformer and uses a patch-splitting layer with the same number of parameters as SpikeCLIP.

ScratchCLIP's text encoder uses an MLP architecture, as well as a word embedding layer of conventional CLIP. 

The detailed training scheme of SpikeCLIP is presented below:
\begin{itemize}
    \item Images size: $32 \times 32$.
    \item Neuron Threshold:
    \begin{itemize}
        \item Spiking neurons of self-attention blocks : $U_{at} = 0.25$;
        \item Other spiking neurons: $U_{thr} = 1.0$.
    \end{itemize}
    \item Decay rate: $\beta = 0.9$.
    \item Time step (of peak input): $T = 4$.
    \item Pre-training image encoder:
    \begin{itemize}
        \item Input dimension: $224 \times 224$. 
        \item Batch size: $196$.
        \item Learning rate: $lr_0 = 5 \times 10^{-3}$ and cosine decay is employed in the first 50 epochs and $lr = 5 \times 10^{-4}$ remain unchanged after the first 50 epochs. The equation is given by:
            \[
                lr(t) = 
                \begin{cases} 
                2.75 \times 10^{-3} + 2.25 \times 10^{-3} \cos\left(\frac{\pi t}{50}\right) & \text{for } 0 \leq t \leq 50 \\
                5 \times 10^{-4} & \text{for } t > 50
                \end{cases}
            \]
        \item Training epochs: $200$. 
    \end{itemize}
    
    \begin{minipage}[t]{0.5\textwidth}
        \item Pre-training text encoder:
        \begin{itemize}
            \item Batch size: $256$.
            \item Learning rate: $lr = 5 \times 10^{-4}$.
            \item Training epochs: $100$. 
            \item Text length: $20$.
        \end{itemize}
    \end{minipage}
    \hfill
    \begin{minipage}[t]{0.45\textwidth}
        \item Fine-tuning:
        \begin{itemize}
            \item Batch size: 196.
            \item Learning rate: $lr = 5 \times 10^{-4}$.
            \item Training epochs: $400$. 
        \end{itemize}
    \end{minipage}
    \item Devices: $2 \times$ 4 NVIDIA GeForce RTX 3090 GPUs.
\end{itemize}

\section{Overview of Datasets Used in the Experiments}
\label{appendix:Datasets}
The datasets employed across the aforementioned experiments are delineated below:
\begin{itemize}
\setlength{\itemsep}{0pt}
\setlength{\parsep}{0pt}
\setlength{\parskip}{0pt}

\item{\textbf{ImageNet-1k}}:
The ImageNet-1k serves as a foundational benchmark in computer vision research, comprising approximately 1.2 million high-resolution color images across 1,000 distinct categories. 
The dataset is commonly partitioned into training, validation, and testing subsets to enable rigorous evaluation of machine learning models. Due to its scale and diversity, ImageNet-1k has become instrumental in the development and assessment of state-of-the-art algorithms.
In addition, this dataset is one of the largest image classification datasets available\citep{ILSVRC15}.

\item{\textbf{CIFAR10}}:
The CIFAR10 serves as a well-established benchmark within the domains of machine learning and computer vision. 
Comprising 60,000 color images with a resolution of 32x32 pixels, the dataset is organized into 10 unique classes. 
With each class containing 6,000 images, the dataset ensures a balanced class distribution. Conventionally, CIFAR10 is partitioned into 50,000 images for training and 10,000 images for testing, thereby providing a consistent framework for evaluating the performance of classification models\citep{krizhevsky2009learning}.

\item{\textbf{CIFAR100}}:
An extension of the CIFAR10 dataset, CIFAR100 is also a prominent benchmark in the fields of machine learning and computer vision. 
While maintaining the same overall count of 60,000 color images at a 32x32 pixel resolution, CIFAR100 expands the class diversity to 100 distinct categories, each represented by 600 images. 
For evaluative purposes, the dataset is typically segmented into 50,000 training images and 10,000 testing images. 
This augmented class variety enhances CIFAR100's utility for conducting more nuanced assessments of classification models\citep{krizhevsky2009learning}.

\item{\textbf{Flower102}}:
The Flower102 dataset is a notable asset within the computer vision landscape, explicitly designed to cater to fine-grained image recognition endeavors. 
The dataset comprises a diverse set of images, capturing 102 different floral species. 
Each category is scrupulously curated to maintain a balanced representation, thereby enabling more sophisticated model evaluations. Due to its focus on capturing subtle variances between closely aligned classes, the Flower102 dataset plays a pivotal role in both refining and benchmarking specialized image classification algorithms\citep{DBLP:conf/icvgip/NilsbackZ08}.

\item{\textbf{Caltech101}}:
As an esteemed benchmark in computer vision research, the Caltech101 dataset encompasses an assemblage of approximately 9,000 color images, categorized into 101 distinct object classes. 
These classes span a diverse array of subjects, including animals, vehicles, and inanimate objects, with a fluctuating number of images allocated to each category. 
Widely employed for a variety of computational tasks, such as object recognition and classification, Caltech101 offers a multifaceted visual dataset for the rigorous evaluation of machine learning model performance\citep{1384978}.

\item{\textbf{OxfordIIIPet}}:
The OxfordIIIPet dataset holds a significant position in the realm of computer vision, particularly in the context of fine-grained classification assignments. 
The dataset comprises visual representations of 37 distinct breeds of cats and dogs, furnishing a nuanced foundation for algorithms engineered to discern subtle visual cues. 
Each breed category is populated with a balanced assortment of images, thereby facilitating the compilation of representative training and testing subsets. 
Owing to its targeted emphasis on the classification of pet breeds, the OxfordIIIPet dataset proves invaluable for fine-tuning models aimed at specialized image recognition tasks\citep{6248092}.

\item{\textbf{STL10}}:
The STL10 dataset is characterized by its collection of color images with a 96x96 pixel resolution, and it includes 10 unique categories that parallel those found in the CIFAR10 dataset. 
It is organized into distinct segments: a labeled set that consists of 5,000 images, an unlabeled set with 100,000 images, and an 8,000-image test set reserved for evaluation. 
This configuration provides a versatile framework for both supervised and unsupervised learning approaches, making it a useful resource for a diverse array of machine-learning applications.

\end{itemize}

To train the text encoder, we curated a dataset \textit{D-text} comprising 115,708 textual entries derived from the labels of 27 datasets used in CLIP's zero-shot evaluation, along with their respective templates.
Consider the CIFAR10 dataset as an example: with its 10 labels and 18 associated templates, 180 distinct text segments are generated for \textit{D-text} (\hyperlink{https://github.com/openai/CLIP}{CLIP}). 
A few templates are illustrated below:

\begin{itemize}
    \item \texttt{A blurry photo of a \{\}.}
    \item \texttt{A black and white photo of a \{\}.}
    \item \texttt{A high-contrast photo of a \{\}.}
    \item \texttt{A photo of a big \{\}.}
\end{itemize}

\section{Pre-training Dataset Sizes and Distributions}
\label{appendix:pretraining-data}

Owing to limitations in acquiring a large dataset of image-text pairs, our SpikeCLIP model was unable to undergo the same pre-training scheme as the original CLIP model. Nonetheless, we posit that with access to adequate training data, SpikeCLIP's performance can be enhanced. To substantiate this hypothesis, we designed a specific experimental setup.

Two metrics are used to quantify the amount of training data: data volume and data distribution. The term data volume refers to the total number of samples utilized during training, while data distribution denotes the level of similarity between the training and evaluation data. Our experiments employ two evaluation datasets: CIFAR10 and ImageNet-1k. We set six different levels of training data volume, ranging from 0k to 100k when evaluating on CIFAR10, and 0k to 50k for ImageNet-1k. Regarding data distribution, we establish three different dataset mixing schemes with varying levels of similarity to CIFAR10 and ImageNet-1k, detailed as follows:

\begin{itemize}
    \item \textbf{Pre-training Data for CIFAR10 evaluation:}
    \begin{itemize}
        \item More similar: \(\frac{1}{3}\) CIFAR10 + \(\frac{1}{3}\) CIFAR100 + \(\frac{1}{3}\) ImageNet-1k;
        \item Less similar: \(\frac{1}{2}\) CIFAR100 + \(\frac{1}{2}\) ImageNet-1k;
        \item No similarity: ImageNet-1k only.
    \end{itemize}
    \item \textbf{Pre-training Data for ImageNet-1k evaluation:}
    \begin{itemize}
        \item More similar: \(\frac{1}{3}\) ImageNet-1k + \(\frac{1}{3}\) CIFAR100 + \(\frac{1}{3}\) CIFAR10;
        \item Less similar: \(\frac{1}{2}\) CIFAR100 + \(\frac{1}{2}\) CIFAR10;
        \item No similarity: CIFAR10 only.
    \end{itemize}
\end{itemize}

\section{Designing More Challenging Multimodal Image Classification Tasks}
\label{appendix:image-cls}

To assess the modal alignment capabilities of SpikeCLIP, we designed two distinct experimental paradigms to evaluate its classification ability. The first approach involved \textit{Unseen Label Set}, using the CIFAR10 dataset as a representative example, each label is replaced by its closest label from the CIFAR100 and ImageNet-1k datasets.

The selection process was facilitated through a specific prompt, termed \textbf{Prompt1}, with the assistance of ChatGPT \citep{openai-2022-chatgpt}. Additionally, we conducted four sub-experiments involving random label replacement at different scales: 20\%, 40\%, 80\%, and 100\%. For the initial three scenarios, predefined random seeds were used, and each was executed in triplicate to record both the \textit{mean} and \textit{variance} of the results.

The second experimental paradigm focused on \textit{Expanded Label Set}. Once again employing the CIFAR10 dataset, we used a separate prompt, \textbf{Prompt2}, to engage ChatGPT in the selection of $N \times 10$ labels that were most dissimilar to the original 10 labels of CIFAR10. This effectively expanded the label set by a factor of $(N+1)$. Subsequently, classification accuracy was evaluated under these modified conditions. The two aforementioned prompts are listed below:
\begin{itemize}
\item \textbf{Prompt1:} The following is the label list L1 for dataset $\text{DS}_1$. Please select the label that is closest to label $x: L_1$.
\item \textbf{Prompt2:} The following are the label lists for dataset $\text{DS}_0, L_0$, and $\text{DS}_2, L_2$. Please select $N$ labels from $L_1$ that are the least similar to the labels in $L_0, L_0, L_2$.
\end{itemize}
{\color{black}
In the above Prompts, $\text{DS}_0 \in$ $\{$CIFAR10, STL10$\}$, $\text{DS}_1 \in$ $\{$CIFAR100, ImageNet-1k$\}$, and $\text{DS}_2 \in$ $\{$CIFAR100$\}$.
}

{
\color{black}
\section{Results of Zero-shot Experiments of SpikeCLIP on 26 Datasets} \label{app:zero26}
we follow the setup of CLIP \cite{radford2021learning} to conduct zero-shot experiments using $26$ datasets.
We exclude ImageNet since it has been used for pre-training.
The results are presented in the Table \ref{tab:zero26}.
\begin{table}[h]
\centering
\small
\caption{
\label{tab:zero26}
\color{black}
We demonstrated the zero-shot generalization ability of ScratchCLIP and SpikeCLP on 26 datasets.
}
\resizebox{\textwidth}{!}{
\begin{tabular}{c|cccccccccccccccccccccccccc}
\hline
Model  & \rotatebox{90}{FER2013} & \rotatebox{90}{STL10} & \rotatebox{90}{EuroSAT} & \rotatebox{90}{RESISC45} & \rotatebox{90}{GTSRB} & \rotatebox{90}{KITTI}  & \rotatebox{90}{Country211}  &  \rotatebox{90}{PCAM} &  \rotatebox{90}{UCF101} &  \rotatebox{90}{Kinetics700}  & \rotatebox{90}{CLEVR}  & \rotatebox{90}{HatefulMemes}  &  \rotatebox{90}{SST} \\ \hline
StratchCLIP  & $47.20$    & $75.69$ & $68.42$   & $55.40$     & $52.95$ & $31.80$  & $2.70$  & $21.58$  & $9.80$  & $7.50$  & $2.18$ & $24.74$  & $36.40$\\
SpikeCLIP   & $45.72$   & $77.79$ & $64.25$   & $51.75$    & $53.42$ & $30.15$  & $1.50$ & $20.95$  & $9.95$  & $8.15$  & $3.95$  & $25.18$  & $35.96$ \\ \hline
\end{tabular}
}
\resizebox{\textwidth}{!}{
\begin{tabular}{c|cccccccccccccccccccccccccc}
\hline
Model & \rotatebox{90}{Food101} & \rotatebox{90}{CIFAR10} & \rotatebox{90}{CIFAR100} & \rotatebox{90}{Birdsnap} & \rotatebox{90}{SUN397} & \rotatebox{90}{Cars} & \rotatebox{90}{Aircraft} & \rotatebox{90}{VOC2007} & \rotatebox{90}{DTD} & \rotatebox{90}{Pets} & \rotatebox{90}{Caltech101} & \rotatebox{90}{Flowers} & \rotatebox{90}{MNIST} \\ \hline
StratchCLIP & $11.50$    & $59.70$    & $27.94$    & $4.80$      & $5.30$    & $37.20$ & $14.60$     & $42.30$    & $35.50$ & $48.60$  & $48.72$      & $8.33$    & $89.75$ \\
SpikeCLIP   & $9.80$     & $58.03$   & $26.66$    & $4.50$      & $5.45$   & $35.80$ & $12.20$     & $42.45$   & $36.80$ & $44.89$ & $48.28$      & $9.02$    & $88.55$ \\ \hline
\end{tabular}
}
\end{table}

}

\section{Comparison of {\color{black} Computing} Energy Consumption }
\label{appendix:ECR}

According to \cite{yao2022attention}, the theoretical {\color{black} Computing} energy consumption of layer $l$ in a SNN can be calculated as: 
\begin{equation} \small
    Energy(l) = E_{AC}\times SOPs(l),
    \label{equation:SNN Energy}
\end{equation}
where SOPs are referred to the number of spike-based accumulate (AC) operations. For classical ANNs, the theoretical energy consumption required by the layer $b$ can be estimated by: 
\begin{equation}  \small
    Energy(b) = E_{MAC}\times FLOPs(b),
    \label{equation:ANN Energy}
\end{equation}
where FLOPs is the floating point operations of $b$, which is the number of multiply-and-accumulate (MAC) operations.
Assuming that the MAC and AC operations are implemented on the 45nm hardware \citep{horowitz20141}, where $E_{MAC} = 4.6pJ$ and $E_{AC} = 0.9pJ$ ($1$J $= 10^{3}$ mJ $=10^{12}$ PJ). 

Thus, the number of synaptic operations at the layer $l$ of an SNN is estimated as: 
\begin{equation}  \small
    SOPs(l) = T \times \gamma \times FLOPs(l),
    \label{equation:convert}
\end{equation}
where $T$ is the number of time steps required in the simulation, $\gamma$ is the firing rate of the input spike train of the layer $l$. 

Therefore, we estimate the theoretical {\color{black} Computing} energy consumption of SpikeCLIP as follows:
\begin{equation} \small
    E_{SpikeCLIP} = E_{AC}\times \left(\sum_{m = 1}^M\text{SOP}_{\text{SNN FC}}^m + \sum_{n = 1}^N\text{SOP}_{\text{SNN Conv}}^n\right),
    \label{equation:SNN Energy Calculation}
\end{equation}
where $\text{SNN FC}$ and $\text{SNN Conv}$ are the fully connected linear layer and the convolutional layer with neurons in SpikeCLIP respectively.
As shown in Equation \ref{equation:SNN Energy Calculation}, the SOPs of $m$ SNN Fully Connected Layer (FC), $n$ SNN Convolutional layers are added together and multiplied by $E_{AC}$.

We refer to \cite{horowitz20141}, assuming that MAC and AC operations are implemented on 45nm hardware (the calculation of power consumption in this hardware only involves MAC and AC operations) since SpikeCLIP and ScratchCLIP have the same architecture except for pulsar neurons, We can calculate the energy consumption reduction (ECR) by equations \ref{equation:SNN Energy}, \ref{equation:ANN Energy}, \ref{equation:convert} and \ref{equation:SNN Energy Calculation} as the following expression equation:
\begin{equation}  \small
    ECR = 1 - \frac{E_{AC} \times T \times \bar{\gamma}}{E_{MAC}},
    \label{equation:ECR}
\end{equation}
where $E_{MAC} = 4.6pJ$, $E_{AC} = 0.9pJ$, and $\bar{\gamma}$ represent the average neuron firing rate of the whole SpikeCLIP.

{\color{black}
\section{Limitations}\label{app:limit}
}

In this study, we have embarked on one of the initial endeavors to employ spiking neural networks (SNNs) in multimodal tasks, with a particular emphasis on classification tasks.
It would indeed be intriguing to broaden the scope of this work to encompass generative tasks, such as image captioning and visual question answering.
We relied on an existing CLIP to pre-train SpikeCLIP by distilling knowledge from the conventional CLIP.

{\color{black}
Another limitation lies in our estimation of energy consumption. While we follow previous studies \citep{yao2022attention, zhou2022spikformer} for calculating energy consumption, their methods only consider computing energy. However, memory access and data movement are also significant factors that affect the overall energy consumption of SNNs, and we further discuss their impact as follows:

\textbf{Memory access plays a crucial role in the energy consumption of SNNs, especially when considering the deployment of SNNs on real hardware platforms like FPGAs or specialized neuromorphic chips, such as Loihi \cite{davies2018loihi}.}
Unlike conventional ANNs, where the energy consumption is primarily driven by the Multiply-and-Accumulate (MAC) operations, SNNs rely heavily on sparse event-driven computations.
While the computational load in SNNs may appear lower due to fewer spikes being processed compared to the dense activations in ANNs, the energy cost associated with memory access is often overlooked.
In SNNs, the energy expenditure is not just due to MAC or Accumulate (AC) operations but also the need to frequently read and write spike data to memory, especially when dealing with large-scale networks.
The memory access patterns in SNNs can result in significant energy overhead, particularly in scenarios where spikes are stored in large buffers or memory arrays.
This becomes even more critical when hardware platforms rely on external memory, where the energy cost of fetching and storing spikes can dominate the overall energy budget. Thus, while SNNs might appear more energy-efficient at first glance due to their sparse nature, the energy consumption tied to memory access can significantly reduce these advantages, especially in hardware with high memory access latencies or limited bandwidth.
This aspect of energy consumption must be carefully considered when evaluating the overall efficiency of SNNs.
}


\newpage











\bibliographystyle{unsrtnat}
\bibliography{nn_2024}
\end{document}